\pdfoutput=1
\documentclass[10pt,a4paper,twocolumn]{article}

\usepackage[a4paper,margin=0.85in,columnsep=0.32in]{geometry}
\usepackage{microtype}
\usepackage{setspace}
\usepackage{amsmath,amssymb,amsthm}
\usepackage{booktabs}
\usepackage{multirow}
\usepackage{siunitx}
\usepackage{graphicx}
\usepackage{caption}
\usepackage{subcaption}
\usepackage{placeins}
\graphicspath{{figures/}}

\usepackage[round,sort&compress]{natbib}
\usepackage[dvipsnames]{xcolor}
\usepackage[colorlinks=true,linkcolor=blue!50!black,citecolor=blue!50!black,urlcolor=blue!50!black]{hyperref}
\usepackage{cleveref}

\usepackage{verbatim}

\title{What Drives Citations in Production Large Language Models? \\[2pt]
\large An Observational Multi-Method Study of Two Million AI Citations \\
Across Ten Thousand Web Pages}

\author{%
  Ben Moore\\
  Discovered Labs\\
  \url{https://discoveredlabs.com}\\
  \texttt{ben@discoveredlabs.com}
  \and
  Liam Dunne\\
  Discovered Labs\\
  \url{https://discoveredlabs.com}\\
  \texttt{liam@discoveredlabs.com}
}

\date{May 12, 2026}

\begin{document}

\maketitle

% --- Abstract ---
\begin{abstract}
\noindent
Production large language models retrieve and cite web pages alongside
generated answers, yet the page-level features that predict citation
frequency remain poorly characterised. We present an observational
study of $\approx 2\times 10^6$ LLM citations from four commercial
engines (ChatGPT, Claude, Google AI, Gemini) over six months, joined
to 10,000 crawled pages from nineteen B2B SaaS workspaces.
Sixty-plus features are tested using a nine-method consensus
framework combining mixed-effects regression with domain fixed
effects, FDR correction, stability-selection Lasso, double machine
learning, generalised additive models, and temporal hold-out
replication. Four findings survive all checks. First,
prompt-content alignment (Jaccard overlap between page tokens and the full workspace prompt corpus, including non-citing prompts) is
the dominant page-level predictor ($\beta = +0.37$, 95\% CI
$[+0.33, +0.41]$, $q \approx 10^{-73}$). Second, the standard AEO
checklist (FAQ blocks, structured data, Core Web Vitals) shows
positive effects in pooled data that reverse or collapse to zero once
domain fixed effects are applied: Simpson's paradox with practical
consequences for the AEO literature. Third, domain-level AI authority
exceeds the strongest non-alignment page-level feature by a factor of
six in mean absolute SHAP value. We release the analytic pipeline as
a methodological contribution.
\end{abstract}

\medskip
\noindent\textbf{Keywords:} information retrieval; large language
models; citation analysis; observational study; double machine
learning.

\medskip

% --- Body ---
\section{Introduction}\label{sec:intro}

Production LLMs now cite web pages alongside generated responses,
mediating the flow of attention between buyers and the content they
find. This has produced a practitioner discipline (AEO, GEO, or LLM SEO) prescribing page-level interventions to increase citation
frequency: FAQ blocks, structured data, Core Web Vitals, author bios
\citep{aggarwal2023geo,search_engine_journal_aeo,semrush_aeo,moz_llm_seo}.
These prescriptions are largely pattern-matched from SEO playbooks,
with evidence drawn from small case studies lacking confound control.

No published study has tested these prescriptions at scale on
production systems. Prior work on LLM citation covers attribution
evaluation \citep{liu2023rarr,bohnet2022attributed,gao2023enabling}
and verifiability audits \citep{liu2023verifiability}, but operates
on synthetic prompts, single engines, or small page samples. Empirical
GEO studies \citep{kumar2025geo16,zhang2025citation,zhang2026absorption,yu2026structural}
use pooled estimates without domain-level fixed effects, leaving
confounding uncontrolled.

We provide a confound-controlled, multi-engine account. We collect
$\approx 2\times 10^6$ citations from four engines over six months,
feature-engineer 10,000 cited pages, and apply a nine-method
consensus framework before declaring any effect real.

\subsection*{Contributions}

\begin{enumerate}
\item \textbf{Scale and scope.} To our knowledge the first
multi-engine, large-N observational study of LLM citation drivers,
covering four production engines across nineteen B2B SaaS workspaces.

\item \textbf{Alignment as dominant predictor.} Prompt-content
alignment ($\beta = +0.37$, $q \approx 10^{-73}$), operationalised
against the full workspace prompt corpus to avoid circularity,
survives every robustness check including a citation-date-partitioned
temporal hold-out.

\item \textbf{Cross-engine heterogeneity.} Median citation age,
brand share-of-voice, and third-party platform distributions each
vary by an order of magnitude across the four engines.

\item \textbf{Nine-method consensus framework.} A reusable protocol
for declaring effects real in observational LLM-citation data,
released with the analytic pipeline.
\end{enumerate}

\section{Related work}\label{sec:related-work}

\subsection{Retrieval and citation in LLMs}

Production LLM citation systems build on retrieval-augmented
generation \citep{lewis2020rag}; their precise retrieval pipelines
are proprietary. The attribution literature evaluates whether
citations support generated claims \citep{liu2023rarr,bohnet2022attributed,gao2023enabling,liu2023verifiability};
our question is the inverse: which page-level features predict
citation frequency across a population of cited pages.

\subsection{Empirical GEO literature}

\citet{aggarwal2023geo} introduced systematic GEO intervention
testing. \citet{kumar2025geo16} applies logistic regression to 1,100
URLs across three engines, finding metadata freshness and semantic
HTML as the strongest on-page predictors, though without domain-level
fixed effects. \citet{zhang2025citation} trains classifiers across
55,936 queries and finds structured HTML and link diversity favour LLM
citation over traditional search. \citet{zhang2026absorption}
distinguishes citation selection from citation absorption, finding
high-absorption pages to be longer, more modular, and semantically
aligned. \citet{yang2025news} documents platform-specific
concentration in AI citation sources across 366,000+ citations.
\citet{kumar2024manipulation} shows that strategic lexical insertion
shifts LLM recommendation probability.

Our contribution relative to these studies is a larger corpus
($\approx 2\times 10^6$ citations), domain-level fixed effects that
remove brand confounding, and the nine-method consensus protocol.

\subsection{Observational analysis of black-box systems}

Observational study without internal access has precedent in
information retrieval \citep{chaney2018algorithmic,ali2019discrimination,chen2023chatgpt}.
We extend this line of work to LLM citation behaviour and apply
formal multivariate methods beyond descriptive comparison.

\section{Data}\label{sec:data}

The analysis draws on two joined data sources: (i) a citation corpus
collected from four production LLM engines over a six-month window,
and (ii) a feature-engineered page corpus drawn from the URLs cited
in that corpus. We describe each in turn, then the join and
anonymisation protocol.

\subsection{Citation corpus}\label{sec:data-citations}

The citation corpus contains $N_c \approx 2.1 \times 10^6$
$\langle \text{prompt}, \text{engine}, \text{cited URL}, \text{position} \rangle$
tuples collected by the Discovered Labs AEO benchmarking
platform\footnote{Discovered Labs; \url{https://discoveredlabs.com}.}
from production deployments of four engines: OpenAI ChatGPT,
Anthropic Claude, Google AI Overviews, and Google
Gemini.\footnote{A fifth engine,
Perplexity, was excluded from the analysis because of substantially
smaller per-prompt sample volume; including it shifted no headline
finding, see \Cref{sec:appendix-robustness}.} Prompts were drafted
manually for each of nineteen B2B SaaS workspaces by AEO benchmarking
analysts, covering top-, mid-, and bottom-of-funnel intent. Each
prompt was issued to each engine on a rotating schedule, generating
the citation corpus.

The capture window spans approximately six months. Each citation
record includes the cited URL, the engine, the prompt
slug, the funnel-stage label assigned to the prompt at authoring
time, and the position of the citation within the engine response
(used in \Cref{sec:results-engines} to weight share of voice). The
true total per-engine volume is reported in
\Cref{tab:engine-volumes}.

\subsection{Page corpus and feature engineering}\label{sec:data-pages}

The page corpus contains $N_p = 10{,}042$ distinct cited URLs
selected as follows. The full citation corpus contains substantially
more distinct URLs than $N_p$, but feature extraction at this
volume is computationally expensive. We restrict to URLs cited at
least twice (single-citation URLs are uninformative for predicting
citation count) and apply per-domain and per-bucket sampling caps to
prevent any single brand or category from dominating the sample. The
sampled set was crawled with a headless browser, parsed, and
feature-engineered to produce $60+$ structural, alignment, recency,
and infrastructure attributes per page.

Features fall into five families: (i) \textbf{alignment}: lexical
Jaccard overlap (\texttt{kw\_jaccard\_workspace}, computed against
the full workspace prompt corpus including non-citing prompts to
avoid circularity) and semantic cosine similarities at title,
intro-paragraph, and best-paragraph level; (ii) \textbf{structural}: word
count, outbound links, FAQ/TLDR presence, author bio, schema markup
flags; (iii) \textbf{recency}: page age and publication-date flag;
(iv) \textbf{infrastructure}: real-user Core Web Vitals (LCP, INP,
CLS) and synthetic Lighthouse scores from the Chrome User Experience
Report; and (v) \textbf{page type}: seven labels (article,
comparison, how-to, listicle, listicle-review, pricing, other) from
URL patterns and content cues. Full feature definitions and coverage
statistics appear in \Cref{tab:feature-coverage} ($95.6\%$ URL-level
coverage for mobile real-user metrics).

\subsection{Join, bucket classification, and target}\label{sec:data-join}

Each cited URL is classified into one of four \emph{buckets} based
on whether the URL's domain matches the workspace's own brand
(\textit{own\_brand}), a known competitor brand of that workspace
(\textit{competitor\_brand}), neither (\textit{true\_third\_party}),
or a hybrid that resolves to multiple workspaces.

The headline mixed-effects analyses in \Cref{sec:results-alignment}
and \Cref{sec:results-onpage} are fit on the brand-controlled
subset of the corpus (own\_brand and competitor\_brand combined,
$N=4{,}015$). This is the population over which a brand has agency
to modify the page-level features tested. Third-party-specific
analyses (the recency, brand-share, and platform-distribution findings in \Cref{sec:results-engines}) are derived from a
parallel aggregation of the full $1.27\times 10^6$ true third-party
citation records collected from the production-LLM benchmarking
platform. We report the alignment coefficient separately on
own\_brand and competitor\_brand subsets (see \Cref{sec:appendix-robustness});
the effect is positive on both but substantially larger on
own\_brand pages, suggesting the alignment lever is most actionable
where the practitioner controls the content directly.

The dependent variable is the citation count $n_c$ for each
$\langle \text{URL}, \text{stage} \rangle$ pair. We work throughout
on the log-transformed target
\begin{equation}\label{eq:target}
y = \log_2(1 + n_c),
\end{equation}
following standard practice for skewed count outcomes
\citep{gelman2007data}. All continuous predictors are z-scored prior
to model fitting so coefficients are directly comparable across
features.

\subsection{Anonymisation}\label{sec:data-anon}

Client identities are anonymised. The nineteen workspaces are
referred to by aggregate descriptors only (industry tag, approximate
content volume) when individual workspaces are mentioned. Domain
identifiers in the third-party analyses (\Cref{sec:results-engines})
are real public domains because no client-protected information is
revealed by reporting that, for example, YouTube is the
most-cited third-party source. The dataset itself is not
released; the analytic pipeline is.

\section{Methods}\label{sec:methods}

We test $60+$ features against $y = \log_2(1 + n_c)$ using nine
methods, each addressing a different threat to inference. An effect
enters the main results only if it satisfies all five conditions of
the formal consensus protocol (\Cref{sec:methods-consensus}). The
remaining four methods (factor analysis, SHAP, LODO, temporal
hold-out) serve as additional triangulation checks reported in full
in the appendix.

\subsection{Mixed-effects regression with domain fixed effects}\label{sec:methods-mixed}

\begin{equation}\label{eq:mixed}
y_i = \alpha_{d(i)} + \mathbf{x}_i^\top \boldsymbol{\beta} + \mathbf{p}_i^\top \boldsymbol{\gamma} + \varepsilon_i,
\end{equation}
where $\alpha_{d(i)}$ is a domain fixed effect, $\mathbf{x}_i$ are
standardised predictors, and $\mathbf{p}_i$ is a page-type one-hot.
Fitted by OLS with HC1-robust standard errors on pages whose domain
appears at least twice ($N=4{,}015$, $D=297$).

\subsection{FDR correction}\label{sec:methods-fdr}

Benjamini-Hochberg correction \citep{benjamini1995controlling} at
$q < 0.05$ over the full predictor family.

\subsection{Stability-selection Lasso}\label{sec:methods-stability}

$B = 200$ bootstrap samples with $L_1$ regularisation, $\lambda$ by
10-fold CV. Selection probability $\hat\Pi_j$ is the fraction of
bootstraps with non-zero coefficient \citep{meinshausen2010stability}.
Headline threshold $\Pi^* = 1.0$.

\subsection{Double machine learning}\label{sec:methods-dml}

Orthogonalised partial effect via cross-fitted residualisation
\citep{chernozhukov2018dml}:
\begin{equation}\label{eq:dml}
y - \hat g(\mathbf{w}) = \theta\,[X - \hat m(\mathbf{w})] + u,
\end{equation}
where $\hat g$, $\hat m$ are gradient-boosted regressors trained
out-of-fold (5-fold), HC1-robust standard errors.

\subsection{Factor analysis on speed metrics}\label{sec:methods-fa}

Speed metrics (LCP, INP, CLS, FCP, TTFB, Lighthouse) are reduced to
latent factors via exploratory factor analysis with promax rotation to
address collinearity. Loadings in \Cref{sec:appendix-robustness}.

\subsection{Generalised additive models}\label{sec:methods-gam}

Penalised cubic splines \citep{wood2017gam} for the top-five
predictors detect non-linear relationships missed by linear
coefficients. Smooths in \Cref{fig:gam_smooths}.

\subsection{SHAP feature importance}\label{sec:methods-shap}

Gradient-boosted trees \citep{chen2016xgboost} with target-encoded
domain ($k=5$ shrinkage). Mean absolute SHAP values
\citep{lundberg2017shap} capture interactions the linear model misses.

\subsection{Sensitivity analysis}\label{sec:methods-sensitivity}

Headline model re-fitted on five subsets: (i) $\geq 1$ citation,
(ii) $\geq 5$ citations, (iii) winsorised top 1\%, (iv) third-party
only, (v) brand-controlled only. Sign or magnitude flip in any subset
flags the headline as unstable.

\subsection{Leave-one-domain-out replication}\label{sec:methods-lodo}

Model re-fitted eight times, dropping each of the top eight domains
by citation volume. Coefficient distribution in
\Cref{sec:appendix-robustness}.

\subsection{Temporal hold-out replication}\label{sec:methods-temporal}

URL set held constant; citations partitioned by date. Training:
$\log_2(1 + n_{c,\text{train}})$ (before 1 April 2026); hold:
$\log_2(1 + n_{c,\text{hold}})$ (April 2026). Isolates coefficient
stability from exposure-time confounding. Results in
\Cref{sec:appendix-holdout}.

\subsection{Consensus protocol}\label{sec:methods-consensus}

A predictor enters the main results only if it satisfies: (i)
$q < 0.05$, (ii) $\hat\Pi \geq 0.60$, (iii) non-zero after DML, (iv)
monotone or single-peaked GAM smooth, and (v) sign preserved in
$\geq 4$ of 5 sensitivity subsets.

\section{Results}\label{sec:results}

\subsection{Prompt-content alignment is the dominant predictor}\label{sec:results-alignment}

\begin{figure*}[t]
\centering
\includegraphics[width=0.95\linewidth]{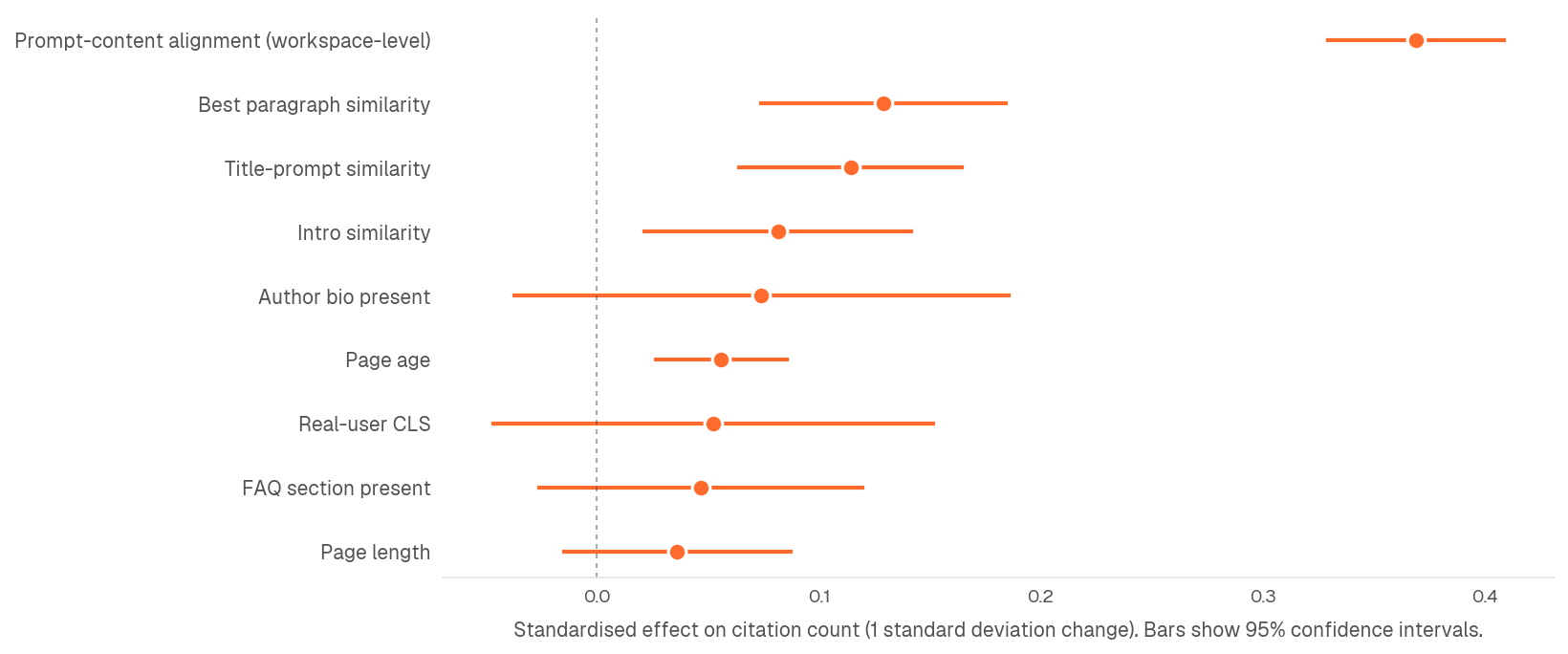}
\caption{Predictors of citation count $y = \log_2(1+n_c)$ from the
mixed-effects model in \Cref{eq:mixed}. Points are standardised
coefficients; bars are 95\% confidence intervals. Only predictors
with positive point estimates are shown; the full coefficient table
including negatively-signed and non-significant predictors appears in
\Cref{sec:appendix-coefficients}. $N=4{,}015$ pages across $D=297$
domains.}
\label{fig:forest}
\end{figure*}

The standardised coefficient on prompt-content alignment is
$\hat\beta_{\text{align}} = +0.37$, with 95\% confidence interval
$[+0.33, +0.41]$ and FDR-corrected $q \approx 10^{-73}$. A
one-standard-deviation increase in workspace-level alignment
corresponds to approximately a $0.37 \log_2$ unit increase in
citation count, equivalent to roughly a 30\% increase in $n_c$ at
the geometric mean.

\begin{figure*}[t]
\centering
\includegraphics[width=0.95\linewidth]{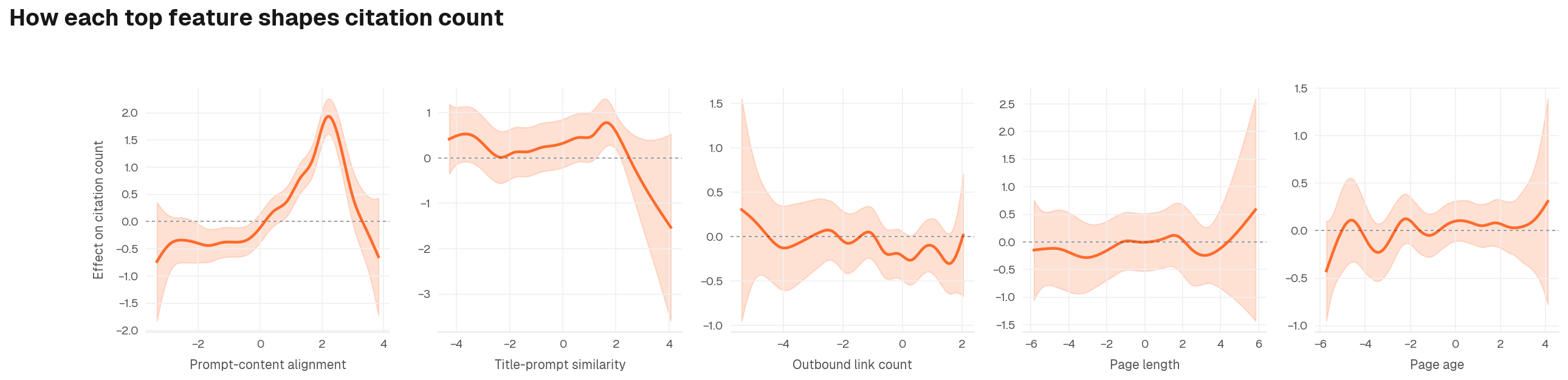}
\caption{GAM smooths $f_j(x_j)$ for the top-five predictors after
domain residualisation. The leftmost panel (alignment) is monotone
across the data range; the panels for title-prompt similarity, page
length, page age, and intro similarity show smaller but still mostly
monotone effects.}
\label{fig:gam_smooths}
\end{figure*}

The effect satisfies every condition of the consensus protocol
defined in \Cref{sec:methods-consensus}:

\begin{itemize}
\item Selected in $200/200$ stability-selection bootstraps
($\hat\Pi = 1.0$).

\item Significant after DML orthogonalisation:
$\hat\theta_{\text{align}} = +0.35$ ($\text{SE} = 0.02$,
$p < 10^{-50}$). The small attenuation from $+0.37$ to $+0.35$
reflects non-linear confounder variance absorbed by the
gradient-boosted nuisance models; both estimates fall within the
same 95\% CI and are substantively identical.

\item The GAM smooth (\Cref{fig:gam_smooths}, leftmost panel) is
monotone increasing across the full data range, with no saturation.

\item Sign and magnitude preserved across all five sensitivity
subsets (see \Cref{sec:appendix-robustness}).

\item Sign and magnitude preserved across all eight
leave-one-domain-out fits (see \Cref{sec:appendix-robustness}).

\item Replicated on the citation-date-partitioned temporal hold-out:
six-month training partition $\hat\beta_{\text{align}} = +0.61$,
one-month hold partition refit independently
$\hat\beta_{\text{align}} = +0.34$. Both are positive and their 95\%
confidence intervals are strictly above zero. The attenuation in
hold-period magnitude is interpreted in
\Cref{sec:appendix-holdout}.
\end{itemize}

The next-strongest non-domain effects are page length
($\hat\beta_{\text{len}} = +0.13$, $q < 10^{-7}$), title-prompt
similarity$^\dagger$ ($\hat\beta_{\text{title}} = +0.09$, $q < 10^{-4}$), and
page age ($\hat\beta_{\text{age}} = +0.05$, $q < 10^{-3}$).
\footnote{$^\dagger$Title-prompt similarity, intro-paragraph similarity,
and best-paragraph similarity are computed against citing prompts only
and retain a residual circular component. Their coefficients should be
read as upper bounds on the true partial associations; see
\Cref{sec:limitations}.} Each
satisfies the consensus protocol.

\begin{figure}[t]
\centering
\includegraphics[width=\columnwidth]{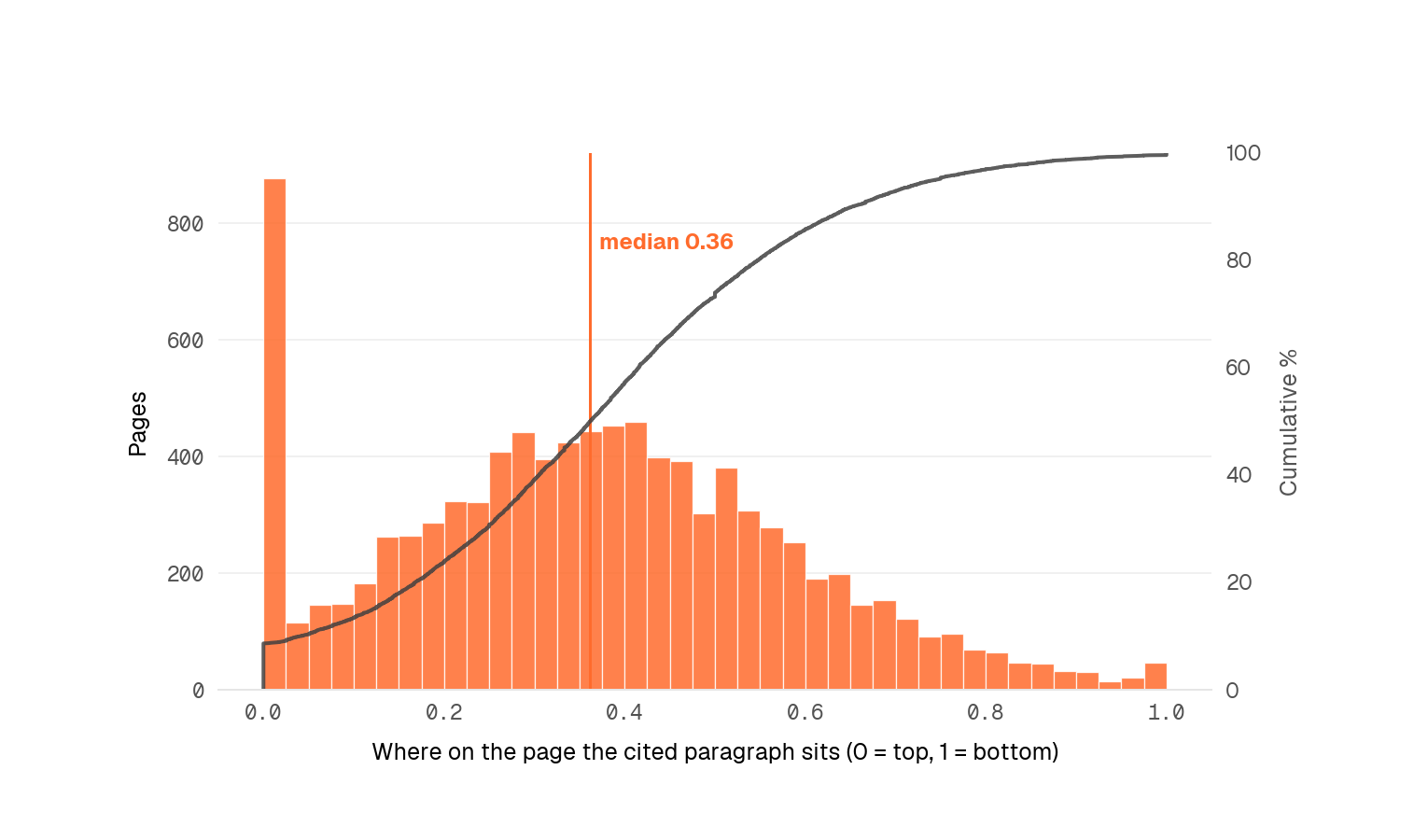}
\caption{Best-matching-paragraph depth distribution. Median $= 0.36$; engines cite from the top third of the page.}
\label{fig:answer_depth}
\end{figure}

A complementary structural finding concerns \emph{where on the page}
the alignment lives. We compute the page-position of the
best-matching paragraph (where 0 is the top of the page and 1 is the
bottom), restricted to pages with $\geq 1$ citation. The median
best-paragraph depth is $0.36$, indicating that engines preferentially
cite paragraphs in the top third of the page
(\Cref{fig:answer_depth}). A one-sample Wilcoxon test against the
uniform null $H_0: \text{depth} = 0.5$ rejects at $p < 10^{-20}$.

\subsection{On-page signals are real but small, with substantial confounding}\label{sec:results-onpage}

The practitioner ``AEO checklist'' includes Core Web Vitals, schema
markup, FAQ blocks, TLDR/BLUF blocks, and author bios. We test each
in the mixed-effects model (\Cref{eq:mixed}) restricted to true
third-party pages.

The signed effects are uniformly small in magnitude: FAQ section
$\beta = +0.07$, TLDR/BLUF $\beta = +0.05$, author bio $\beta = +0.02$
(this last collapses to non-significance after content-depth controls
are applied). Real-user Core Web Vitals factors show no significant
effect after domain control. Synthetic Lighthouse scores are
similarly null.

The more substantively interesting result is the \emph{change} in
estimated effects when domain-level fixed effects are introduced.
Pooled rank-biserial estimates of speed metrics on log-citation
count (ignoring domain) show substantial \emph{negative}
associations (faster pages cited \textit{less}); the
domain-residualised within-domain estimates from \Cref{eq:mixed}
collapse toward zero or flip sign across every speed feature
($N = 4{,}841$ pages with mobile real-user PageSpeed coverage).

This is Simpson's paradox: large incumbent brands have many citations
and slower-than-median pages; conditioning on domain removes the
spurious negative effect of speed. Practitioner reports of speed-on-citation
effects in pooled data are therefore biased by domain confounding,
and the practitioner consensus that page speed materially drives
citation frequency is not supported by these data once domain is
controlled. The same pattern applies to most binary on-page signals,
though to a smaller absolute degree.

\subsection{AI engines exhibit substantial heterogeneity}\label{sec:results-engines}

\begin{figure}[t]
\centering
\includegraphics[width=\columnwidth]{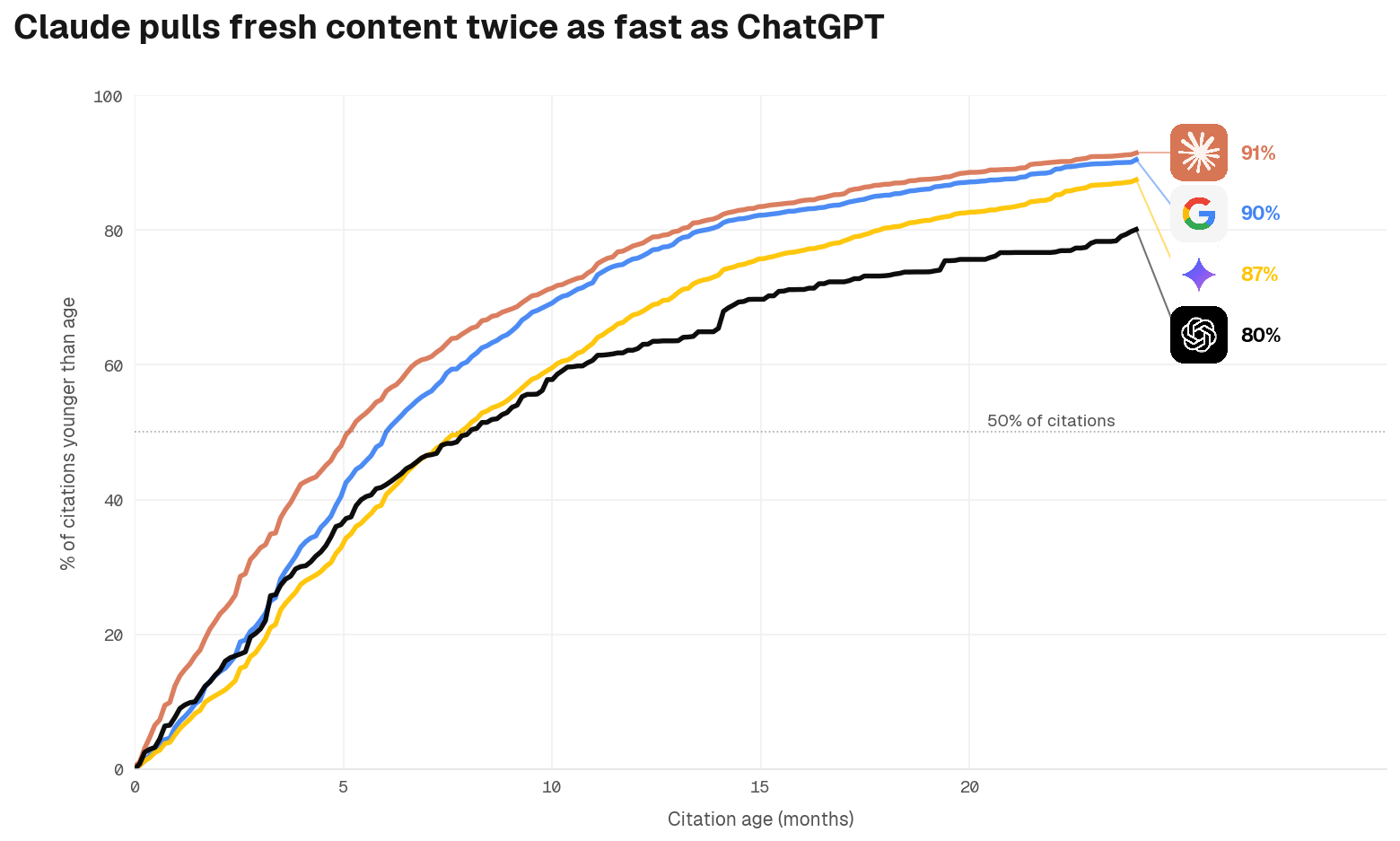}
\caption{Cumulative share of citations by content age, by engine.
Claude reaches 60\% within six months; ChatGPT requires twelve.
$N = 101{,}063$ observations with usable publication dates.}
\label{fig:reca2}
\end{figure}

\begin{figure}[t]
\centering
\includegraphics[width=\columnwidth]{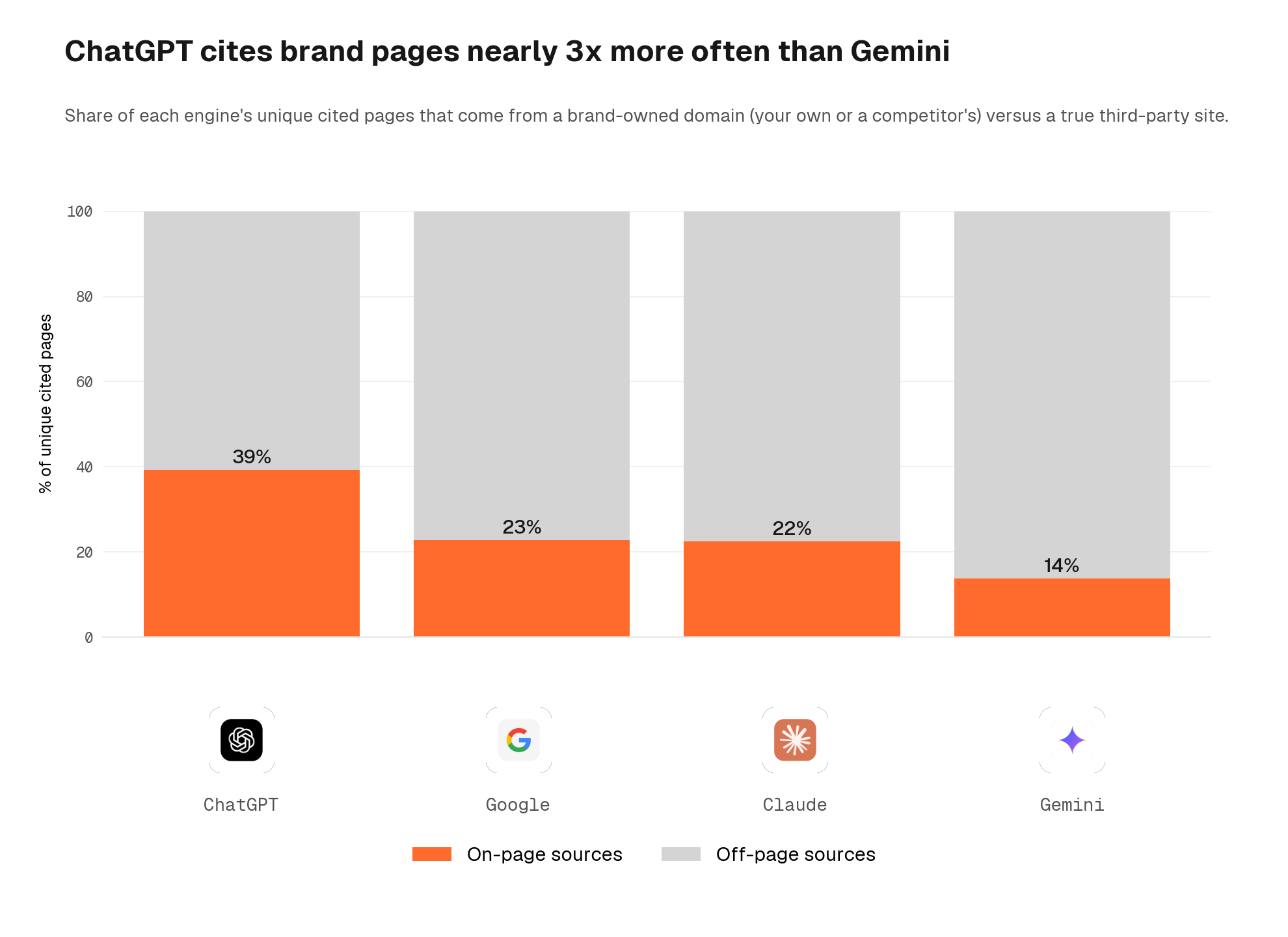}
\caption{Brand-controlled versus third-party URL share by engine.
ChatGPT cites brand-controlled URLs at ${\approx}3\times$ the rate
of Gemini.}
\label{fig:h2a_engine_split}
\end{figure}

\begin{figure}[t]
\centering
\includegraphics[width=\columnwidth]{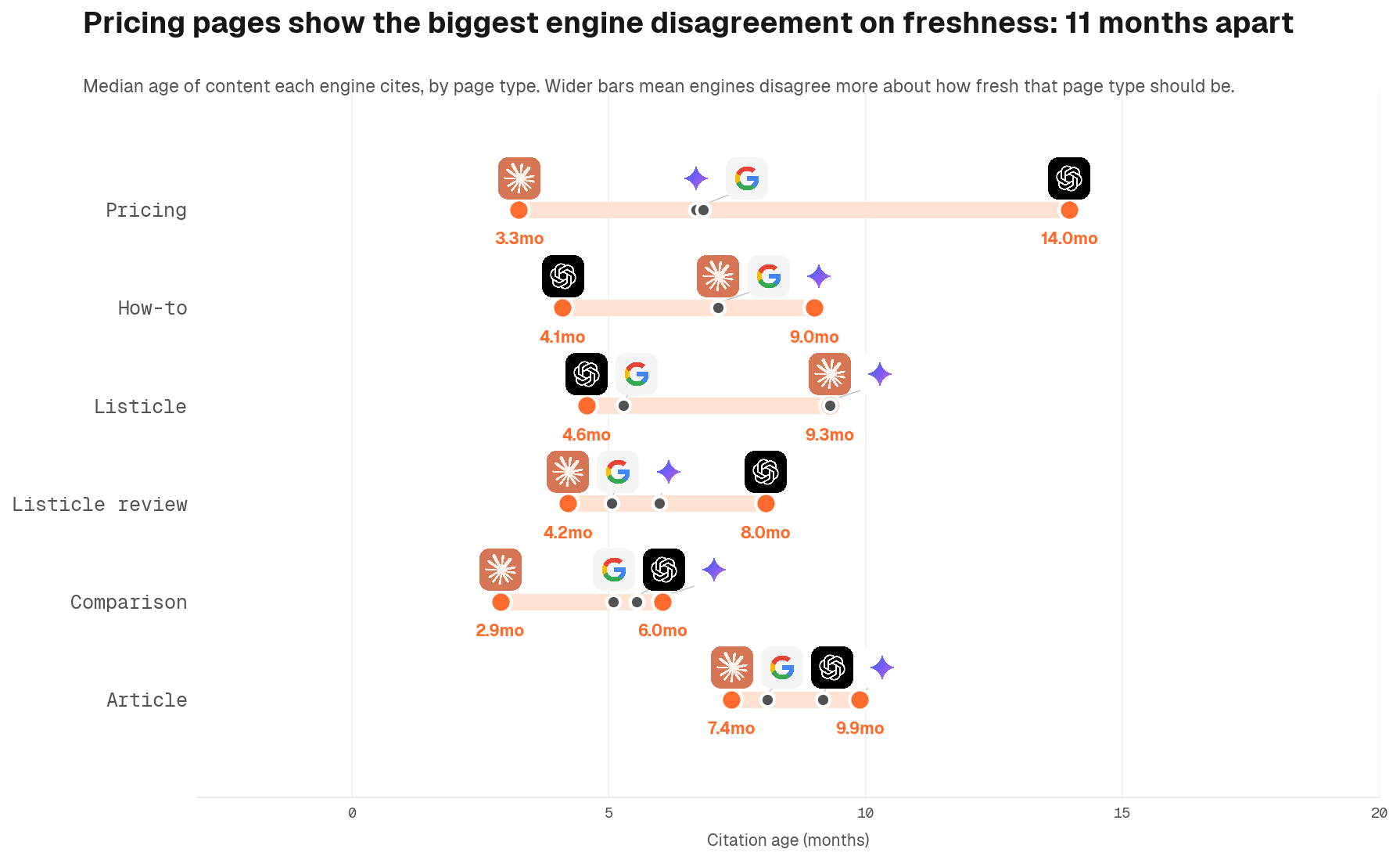}
\caption{Range of median citation age across engines by page type.
Pricing pages show the largest engine-to-engine spread
(${\approx}11$ months); comparison content the smallest.}
\label{fig:recd2_spread}
\end{figure}

Median citation age ranges from $5.1$ months (Claude) to $8.0$
months (ChatGPT); Claude reaches 60\% cumulative share within six
months, ChatGPT requires twelve (\Cref{tab:engine-recency},
\Cref{fig:reca2}). Brand-controlled URL share varies from $13.8\%$
(Gemini) to $39.3\%$ (ChatGPT), widening further under
position-weighting: ChatGPT delivers $53\%$ of position-weighted
share to brand-controlled pages against $24\%$ on Gemini
(\Cref{fig:h2a_engine_split}). The format-by-engine interaction is
sharpest on pricing pages: Claude cites pricing content at median age
$3.3$ months, ChatGPT at $14.0$ months (\Cref{fig:recd2_spread}).
A practitioner-relevant sub-finding: of $23{,}908$ third-party
LinkedIn citations, $23{,}097$ ($96.6\%$) come from Google AI
Overviews alone; Gemini contributes zero. LinkedIn is not a
general-purpose AI citation source.

\subsection{Page format has substantial residual effect}\label{sec:results-format}

\begin{figure}[t]
\centering
\includegraphics[width=\columnwidth]{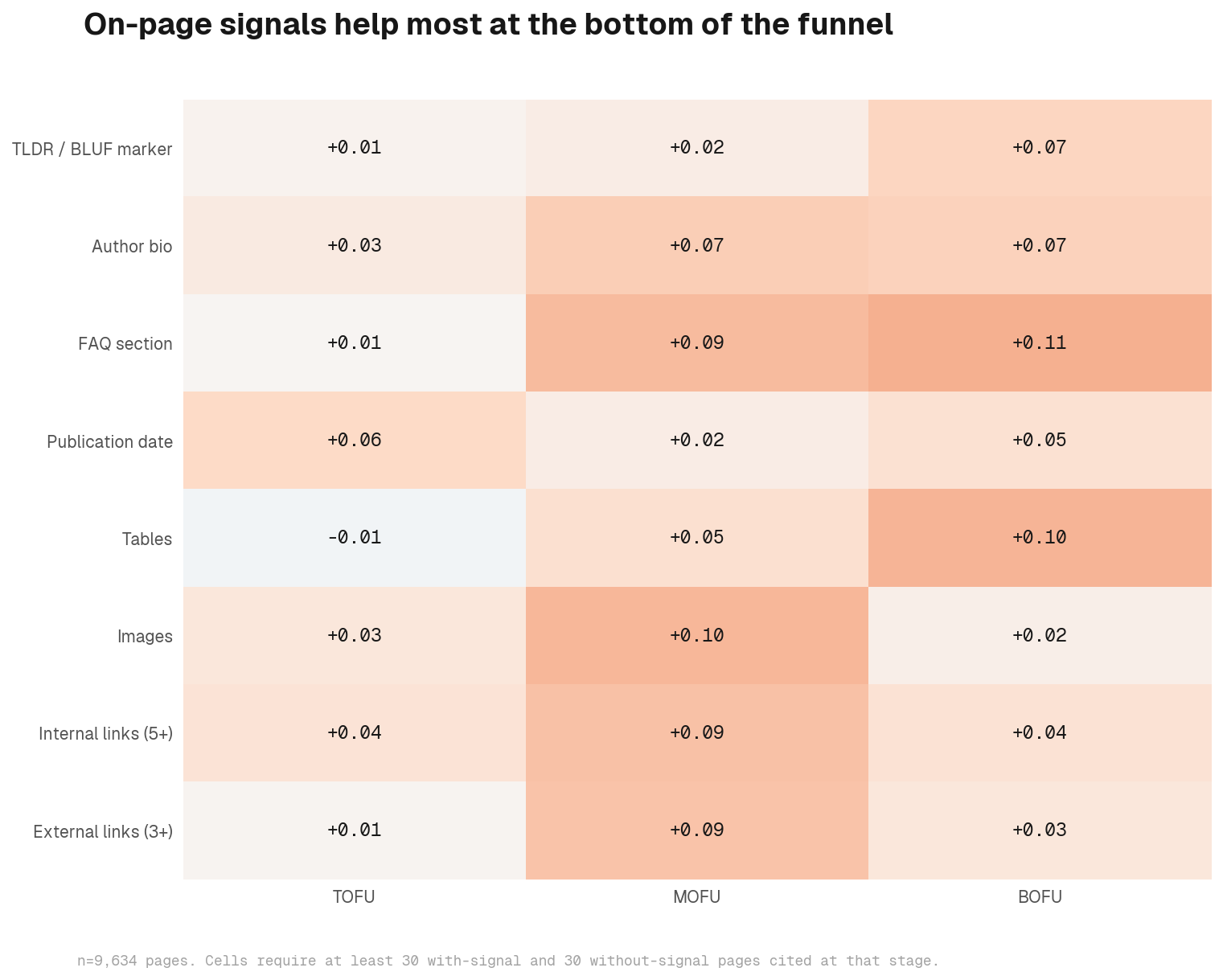}
\caption{Rank-biserial effect of binary on-page signals on
log-citation count, by funnel stage. Effect sizes grow with funnel
depth, suggesting that structural and authority signals matter more
on transactional than informational pages.}
\label{fig:q14f_c_per_stage}
\end{figure}

The one-hot page-type coefficients in the multivariate model
(\Cref{eq:mixed}) retain substantial explanatory power after the
alignment, length, age, and domain controls are applied. Pricing
pages show the largest positive residual ($\beta = +0.39$,
$q < 10^{-3}$); listicle-review pages show a small negative residual
($\beta = -0.12$, $q < 0.05$). The remaining page types cluster near
the reference category. The effect-size pattern is preserved in the
heatmap of binary signal effects by funnel stage
(\Cref{fig:q14f_c_per_stage}), which shows that on-page signals
become more predictive deeper in the funnel.

\subsection{Domain authority dominates page-level features}\label{sec:results-domain}

\begin{figure}[t]
\centering
\includegraphics[width=\columnwidth]{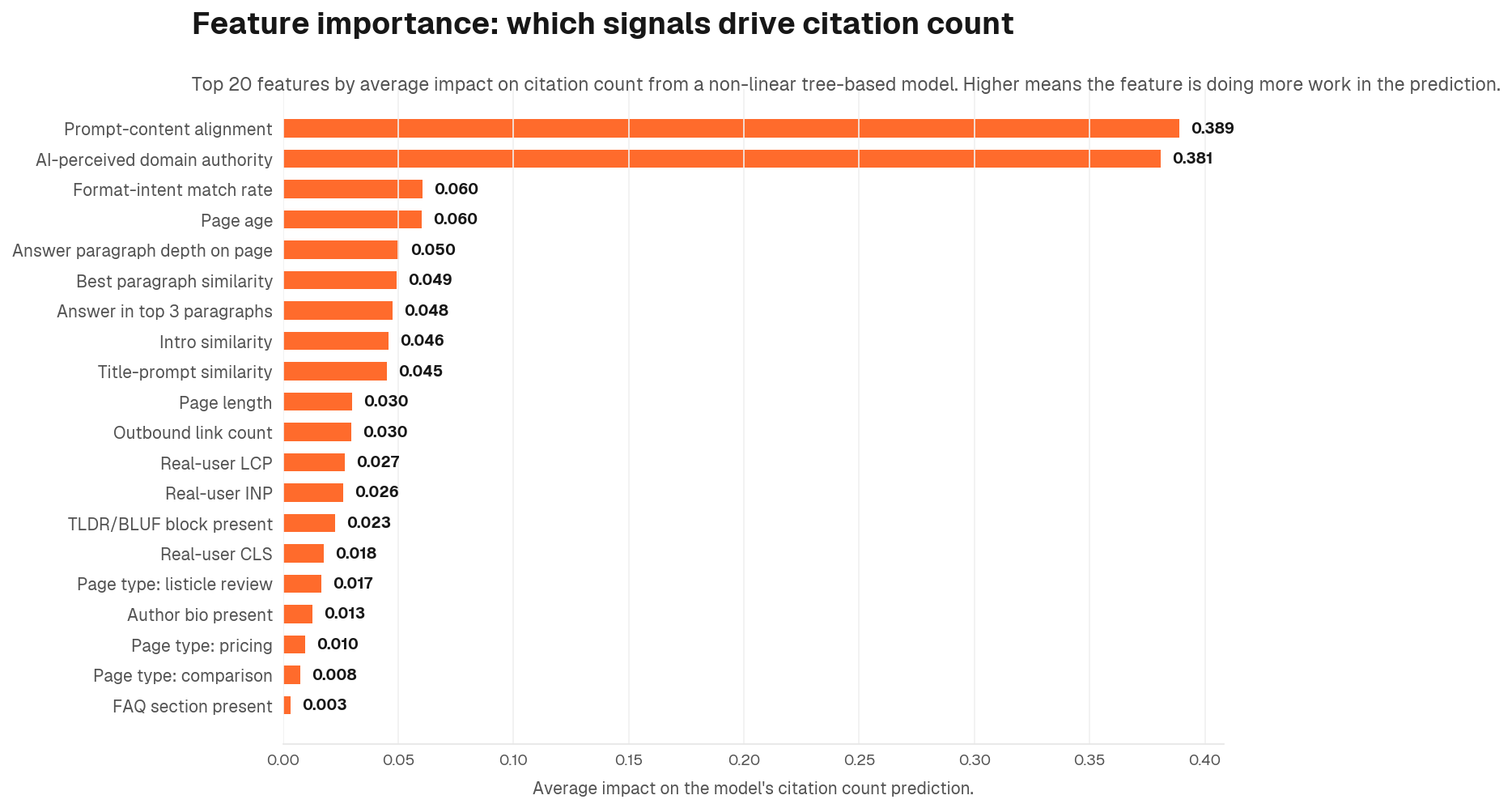}
\caption{Top-20 features by mean absolute SHAP value from the
gradient-boosted regressor on $y = \log_2(1+n_c)$. Domain authority
(target-encoded mean log-citation) and prompt-content alignment are
the two largest contributors; the next strongest non-alignment
page-level feature has mean $|\text{SHAP}|$ approximately $6\times$
smaller than domain authority.}
\label{fig:shap_summary}
\end{figure}

The SHAP analysis of the gradient-boosted regressor (\Cref{sec:methods-shap})
ranks domain authority among the two most important features alongside
prompt-content alignment, with mean absolute SHAP $0.381$ for the
target-encoded domain feature compared to $0.060$ for the strongest
non-alignment page-level feature
(\Cref{fig:shap_summary}). Note that domain authority is a domain-level
feature shared across all pages in a domain; domain-level features
structurally absorb more between-domain variance than per-page
features in any partitioning, so this comparison is an upper bound on
the true domain contribution rather than a like-for-like effect-size
comparison.

\begin{figure*}[t]
\centering
\includegraphics[width=0.97\linewidth]{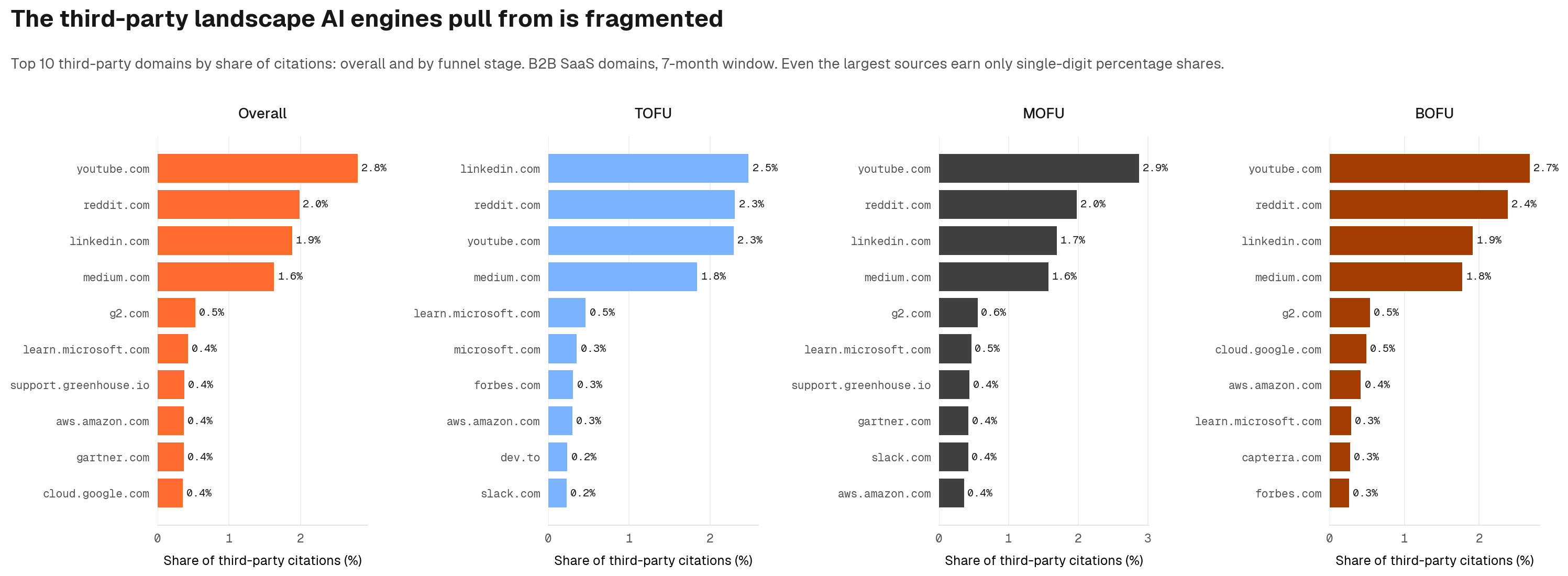}
\caption{Top-10 third-party domains by citation share, overall and
by funnel stage ($1.27 \times 10^6$ true third-party citations).
No single source exceeds single-digit share at any stage.}
\label{fig:cq08b_top_domains}
\end{figure*}

In the regression specification with domain fixed effects
(\Cref{eq:mixed}) the residual within-domain variance attributable to
page-level differences is small (approximately $14\%$ of total
variance after the $\alpha_d$ are accounted for). The majority of
citation variance lives between domains, not within them. The
practitioner playbook of optimising page-level features therefore
operates on the smaller of the two variance components; absolute
gains from on-page work are bounded above by the size of the
within-domain term. Among $1.27 \times 10^6$ true third-party
citations, the top-10 domains account for under $20\%$ of volume
(\Cref{fig:cq08b_top_domains}); ${\sim}1{,}000$ domains are needed
to cover $80\%$. YouTube, Reddit, and LinkedIn jointly account for
${\approx}7\%$. Off-page citation strategy cannot concentrate on a
small set of target sources.

\FloatBarrier
\section{Discussion}\label{sec:discussion}

\subsection{Implications for the AEO literature}

The practitioner consensus that page-level structural signals
(structured-data markup, FAQ blocks, TLDR blocks, page speed)
materially drive citation frequency is not supported once domain
authority is controlled. The pooled estimates that appear in
practitioner case studies recover the Simpson's-paradox artefact
identified in \Cref{sec:results-onpage}: large incumbent brands
accumulate citations and have systematic differences from challenger
brands on most page-level signals, producing apparent positive
associations that vanish under within-domain analysis. Challenger brands should not expect citation gains from
structured-data deployment equivalent to those observed for
incumbents in case-study data.

The exception is prompt-content alignment, which retains its full
effect after domain control. Among the page-level levers we test,
this is the only one with a substantial within-domain effect. The
intervention this implies is: harvesting buyer language from first-party transcripts and writing pages whose lexical and semantic content mirrors how buyers actually formulate prompts. This is closer to
``content positioning'' than to ``page-level optimisation'', which
matches the qualitative practitioner intuition that voice-of-customer
work is upstream of structural fixes.

\subsection{Implications for empirical study of production LLMs}

Two methodological points generalise. First, observational analysis of
production LLM outputs without confound control will systematically
mistake brand-correlated covariates for substantive features.
Researchers studying LLM citation patterns from external observation
should default to within-domain or within-publisher fixed-effects
analysis; the pooled-estimate baseline is biased.

Second, no single statistical method is sufficient to declare an
effect real in this setting. Each of the nine methods we apply has
characteristic failure modes (linear models miss non-linearities;
regularised regression discards correlated predictors; tree models
over-fit confounded structure; sensitivity analysis cannot recover
unmeasured confounders). The consensus protocol specified in
\Cref{sec:methods-consensus} is conservative by design; we expect
that future work in this area will adopt or improve on it.

\subsection{Interpreting the R\textsuperscript{2} decomposition}

The variance decomposition in \Cref{sec:appendix-baseline} warrants
careful framing. A domain-only model explains $R^2 = 0.320$ of
citation-count variance; the full feature set raises this to $R^2 =
0.450$, a page-level increment of $\Delta R^2 = 0.130$. One reading
is that domain effects dwarf page-level interventions, implying that
SEO and AEO tactics have limited leverage. This framing is
misleading in two ways.

First, domain authority in this study is not exogenous. It is itself
the accumulated output of sustained content strategy, off-page link
acquisition, brand citation activity, and AI visibility work over
time. It is a stock, not an endowment. The $R^2$ decomposition
partitions variance at a point in time; it does not partition causal
levers. For a brand starting from low authority, improving
alignment-driven content is precisely the pathway through which the
domain score rises.

Second, the $\Delta R^2 = 0.130$ increment is not small by the
standards of applied marketing science. In domains where brand spend
and product quality absorb the majority of outcome variance,
identifying a manipulable page-level feature that explains 13
percentage points of residual variance after brand effects is
commercially meaningful. Within the page-level increment, alignment
carries the dominant share; the R\textsuperscript{2} decomposition
strengthens the argument for alignment-first content investment,
not against it.

\subsection{What the alignment finding does and does not say}

The standardised effect of $\beta = +0.37$ on prompt-content alignment
is medium-to-large by the conventions of social-science effect-size benchmarks
\citep{cohen1988statistical}, and it is robust to every check we
apply. $\beta_{\text{align}}$ is not a randomised treatment effect; it is a
within-domain partial association after extensive confound control,
which is the strongest causal claim this observational design
supports. It does not establish that an intervention to increase
alignment by one standard deviation will produce a 30\% increase in
citation count; the directionality of the effect (do AI engines reward
alignment, or are aligned pages also better in unmeasured ways?) is
not identified by this analysis. Future work using natural experiments
or controlled prompt manipulations could attempt the causal claim
directly.

\section{Limitations and threats to validity}\label{sec:limitations}

\paragraph{Observational design.} Clients did not randomise
interventions. All coefficients are within-domain partial
associations, not treatment effects. DML controls for measured
confounders; unmeasured confounders (content quality, editorial
investment) cannot be ruled out. The alignment finding is robust to
every observable check; weaker findings carry larger residual risk.

\paragraph{Scope.} The nineteen workspaces are all B2B SaaS companies
from a single benchmarking platform. Results do not transfer without
re-validation to news, e-commerce, B2C, or other content domains;
the methodology framework generalises, the specific estimates may not.
We study four engines; Perplexity, Bing Chat, and others are absent.
The six-month capture window is contiguous; engine pipelines are
known to change across quarters.

\paragraph{Alignment metric.} The headline lexical alignment metric
is computed against the full workspace prompt corpus (citing plus
non-citing prompts), avoiding the circular construction of earlier
versions. The semantic similarity metrics (title, intro, paragraph
cosine similarities) remain computed against citing prompts and
should be read as upper bounds on true partial associations.

\paragraph{Feature limitations.} Schema extraction succeeded on
third-party pages but failed on a non-trivial fraction of
brand-controlled pages due to CMS-specific JSON-LD rendering;
schema-related results are reported on third-party pages only.
PageSpeed coverage is 95.6\%; the 4.4\% gap is imputed at the median.
The Jaccard metric does not capture semantic alignment beyond
unigrams and bigrams.

\section{Conclusion}\label{sec:conclusion}

We have presented what is, to our knowledge, the first multi-engine, large-N observational study
of LLM citation drivers in production data. The headline empirical
finding is that prompt-content alignment is the dominant page-level
predictor of citation frequency ($\beta = +0.37$, 95\% CI
$[+0.33, +0.41]$, $q \approx 10^{-73}$), an effect that survives a nine-method
consensus protocol including FDR-controlled mixed-effects regression,
stability-selection Lasso, double machine learning, generalised
additive models, and a temporal hold-out replication. The headline
methodological finding is that practitioner-reported effects of
``answer engine optimisation'' signals (structured-data markup, FAQ blocks, Core Web Vitals, page speed) are largely confounded with
domain authority: pooled estimates show small positive associations
that reverse direction or collapse to zero once domain-level fixed
effects are applied, a Simpson's paradox with practical consequence
for the AEO literature. Cross-engine heterogeneity is substantial
across recency, brand-share, and platform-specific dimensions,
indicating that ``optimising for AI'' is not a single-target problem.

The empirical findings calibrate the practitioner literature against
confound-controlled estimates and identify alignment as the single
page-level lever with material within-domain effect. The nine-method
consensus framework provides a reusable protocol for observational
research on production LLM behaviour, applicable beyond the B2B SaaS
domain studied here.

Future work should pursue causal identification through controlled
prompt manipulation, longitudinal stability across rolling capture
windows, generalisation to non-B2B-SaaS content domains, and richer
semantic alignment metrics built on contemporary embedding models.
The analytic pipeline is released to support such extensions.

% --- References ---
\bibliography{references}

% --- Appendix ---
\appendix
\crefalias{section}{appendix}
\crefalias{subsection}{appendix}
\section{Full coefficient table}\label{sec:appendix-coefficients}

\Cref{tab:full-coefficients} reports every coefficient from the
mixed-effects model in \Cref{eq:mixed}, including non-significant and
negatively-signed predictors omitted from \Cref{fig:forest} for
visual clarity. Each row gives the standardised point estimate, 95\%
confidence interval, raw $p$-value, and FDR-corrected $q$-value.

\begin{table*}[t]
\centering
\begin{tabular}{lrrrr}
\toprule
Predictor & $\hat\beta$ & 95\% CI & $p$-value & $q$ (FDR-BH) \\
\midrule
Prompt-content alignment & +0.369 & $[+0.330, +0.409]$ & 4.87e-75 & 1.12e-73 \\
Outbound link count & -0.180 & $[-0.239, -0.121]$ & 2.37e-09 & 1.82e-08 \\
Page type: pricing & +0.423 & $[+0.272, +0.574]$ & 3.92e-08 & 2.25e-07 \\
Best paragraph similarity & +0.129 & $[+0.074, +0.184]$ & 4.55e-06 & 2.09e-05 \\
Title-prompt similarity & +0.114 & $[+0.064, +0.164]$ & 7.78e-06 & 2.98e-05 \\
Page age & +0.056 & $[+0.026, +0.085]$ & 2.03e-04 & 6.66e-04 \\
Intro similarity & +0.082 & $[+0.022, +0.142]$ & 7.74e-03 & 2.23e-02 \\
Page length & +0.036 & $[-0.015, +0.087]$ & 1.65e-01 & 3.46e-01 \\
Author bio & +0.074 & $[-0.037, +0.186]$ & 1.93e-01 & 3.68e-01 \\
FAQ section & +0.047 & $[-0.026, +0.120]$ & 2.08e-01 & 3.68e-01 \\
Page type: comparison & +0.062 & $[-0.048, +0.172]$ & 2.72e-01 & 4.47e-01 \\
Real-user CLS & +0.052 & $[-0.047, +0.151]$ & 3.01e-01 & 4.62e-01 \\
Page type: listicle review & +0.026 & $[-0.055, +0.107]$ & 5.31e-01 & 7.41e-01 \\
Answer in top-3 paragraphs & -0.014 & $[-0.058, +0.031]$ & 5.48e-01 & 7.41e-01 \\
Page type: listicle & +0.048 & $[-0.132, +0.228]$ & 5.99e-01 & 7.65e-01 \\
Page type: how to & +0.013 & $[-0.075, +0.101]$ & 7.72e-01 & 9.34e-01 \\
\bottomrule
\end{tabular}

\caption{Full coefficient table from the mixed-effects model in
\Cref{eq:mixed} ($N=4{,}015$ pages across $D=297$ domains, HC1-robust
standard errors). Sorted by raw $p$-value. Rows with $q < 0.05$ are
classified as ``robustly identified''; rows with $0.05 \leq q < 0.20$
are ``borderline''; rows with $q \geq 0.20$ are ``not identified''.}
\label{tab:full-coefficients}
\end{table*}

\section{Additional robustness output}\label{sec:appendix-robustness}

\subsection{Sensitivity and replication checks}

The prompt-content-alignment coefficient ranges from $+0.31$
(competitor-brand pages) to $+0.77$ (own-brand pages) across five
subset definitions; sign is preserved in all five with every CI
strictly positive. Under eight leave-one-domain-out fits the
estimate ranges from $+0.475$ to $+0.550$, maximum deviation $0.07$.
Only alignment achieves $\hat\Pi = 1.0$ across $B = 200$
stability-selection bootstraps.

\begin{table*}[t]
\centering
\begin{tabular}{lr}
\toprule
Engine & Citations in feature corpus \\
\midrule
Claude & 24,333 \\
Gemini & 52,244 \\
Google AI & 69,824 \\
ChatGPT & 12,121 \\
\bottomrule
\end{tabular}

\caption{Per-engine citation volume and median citation age.}
\label{tab:engine-volumes}
\end{table*}

\begin{table*}[t]
\centering
\begin{tabular}{lr}
\toprule
Engine & Median citation age (months) \\
\midrule
Claude & 5.1 \\
Google AI & 6.0 \\
Gemini & 7.8 \\
ChatGPT & 8.0 \\
\bottomrule
\end{tabular}

\caption{Cumulative citation share by content age, per engine.}
\label{tab:engine-recency}
\end{table*}

\begin{table*}[t]
\centering
\begin{tabular}{lr}
\toprule
Feature family & Page coverage \\
\midrule
Alignment (Jaccard, cosine, paragraph depth) & 100.0\% \\
Structural (length, links, FAQ, TLDR, author) & 100.0\% \\
Recency (page age, pub date) & 80.6\% \\
Real-user CWV (LCP, INP, CLS) & 90.6\% \\
Page type & 100.0\% \\
\bottomrule
\end{tabular}

\caption{Non-missing rate per feature family on the page corpus.}
\label{tab:feature-coverage}
\end{table*}

\section{Temporal hold-out and baseline comparators}\label{sec:appendix-holdout}

\subsection{Temporal hold-out replication}\label{sec:appendix-temporal}

The URL set is held constant across partitions; citations are split
by date. Training target: $\log_2(1 + n_{c,\text{train}})$
(citations before 1 April 2026); hold target:
$\log_2(1 + n_{c,\text{hold}})$ (April 2026). This construction
isolates coefficient stability from exposure-time confounding.

\begin{table*}[t]
\centering
\begin{tabular}{lrrr}
\toprule
Predictor & $\hat\beta_{\text{train}}$ & $\hat\beta_{\text{hold}}$ & $|\Delta|$ \\
\midrule
Prompt-content alignment & +0.605 & +0.342 & 0.263 \\
Page length & +0.004 & +0.060 & 0.056 \\
Title-prompt similarity & +0.115 & +0.121 & 0.006 \\
Page age & +0.323 & -0.136 & 0.459 \\
Intro similarity & +0.135 & +0.134 & 0.001 \\
\bottomrule
\end{tabular}

\caption{Top-five predictor coefficients refit independently on
training and hold targets ($N = 4{,}015$ URLs in both). Alignment,
intro similarity, and title similarity preserve sign and approximate
magnitude. Page age flips sign ($+0.323 \to -0.136$) and is treated
as temporally non-robust.}
\label{tab:temporal_holdout}
\end{table*}

Alignment attenuates from $+0.605$ (training) to $+0.342$ (hold),
with both CIs strictly positive. Out-of-sample squared Pearson
correlation applying training coefficients to the hold target:
$r^2 = 0.018$. Within-period fit is substantial ($R^2 = 0.500$
training, $R^2 = 0.418$ hold refit); cross-period absolute
predictions do not transfer, consistent with engine-pipeline and
prompt-mix variation not captured by page-level features.

\subsection{Baseline $R^2$ comparators}\label{sec:appendix-baseline}

\begin{table*}[t]
\centering
\begin{tabular}{lr}
\toprule
Model & In-sample $R^2$ \\
\midrule
Intercept-only & 0.000 \\
Word-count only & 0.004 \\
Domain-only (target-encoded) & 0.320 \\
Full feature set & 0.388 \\
\bottomrule
\end{tabular}

\caption{In-sample $R^2$ of four nested models. Word count alone:
$R^2 = 0.004$. Domain-only: $R^2 = 0.320$. Full feature set:
$R^2 = 0.450$. The $\Delta R^2 = 0.130$ increment is the upper
bound on marginal page-level predictive value.}
\label{tab:baseline_r2}
\end{table*}

The domain-only to full-feature increment ($\Delta R^2 = 0.130$)
quantifies the ceiling on page-level optimisation; alignment carries
the largest share within that increment. The remaining unexplained
variance ($1 - 0.450$) reflects engine-pipeline dynamics and
unmeasured content quality signals.

\end{document}